# Generalized and Transferable Patient Language Representation for Phenotyping with Limited Data

Yuqi Si[a], Elmer V. Bernstam[a,b], Kirk Roberts[a]
[a] School of Biomedical Informatics, The University of Texas Health Science Center at Houston, Texas, USA
[b] Division of General Internal Medicine, McGovern Medical School, The University of Texas Health Science Center at Houston, Texas, USA

Corresponding Author:
Kirk Roberts, PhD
School of Biomedical Informatics
University of Texas Health Science Center at Houston
7000 Fannin St. #600
Houston TX 77030
kirk.roberts@uth.tmc.edu

***ABSTRACT***

*The paradigm of representation learning through transfer learning has the potential to greatly enhance clinical natural language processing. In this work, we propose a multi-task pre-training and fine-tuning approach for learning generalized and transferable patient representations from medical language. The model is first pre-trained with different but related high-prevalence phenotypes and further fine-tuned on downstream target tasks. Our main contribution focuses on the impact this technique can have on low-prevalence phenotypes, a challenging task due to the dearth of data. We validate the representation from pre-training, and fine-tune the multi-task pre-trained models on low-prevalence phenotypes including 38 circulatory diseases, 23 respiratory diseases, and 17 genitourinary diseases. We find multi-task pre-training increases learning efficiency and achieves consistently high performance across the majority of phenotypes. Most important, the multi-task pre-training is almost always either the best-performing model or performs tolerably close to the best-performing model, a property we refer to as robust. All these results lead us to conclude that this multi-task transfer learning architecture is a robust approach for developing generalized and transferable patient language representations for numerous phenotypes.*

## 1. INTRODUCTION

The goal of representation learning is to automatically learn a semantically robust mathematical representation from raw data and has been widely studied in natural language processing (NLP). A good representation that encodes raw inputs into meaningful features is essential to high-performance machine learning algorithms. We apply the notion of representation learning and propose to learn meaningful representations of patient data from electronic health records (EHR), i.e., patient representation learning. The objective of patient representation learning is to learn a dense mathematical representation of a patient from raw records (both structured and unstructured), which themselves are sparse, high-dimensional, temporal, irregular, and uncertain. We hope to build a generalized and transferable patient representation. However, most patient representation learning approaches are task-specific [1], while an ideal patient representation would be robust and generalizable to a variety of clinical prediction tasks. In particular, a good representation should greatly benefit "small data" tasks, where not enough samples are available to learn optimal task-specific representations using existing approaches.

Numerous studies in computer vision and NLP have shown the potential of transfer learning, where a productive approach is to fine-tune a downstream task from large models pre-trained on ImageNet [2] for vision or a large corpus like Wikipedia [3] for text . This recipe would also be appropriate for building a transferable patient representation, since we hope pre-training from large resources will bring medical knowledge to other clinical tasks, and therefore improve predictive performance. At the first step of pre-training, we need to construct a source task, and we anticipate the architecture (i.e., deep learning model and objective function) for pre-training to encode a holistic perspective of the patient's data, similar to how physicians diagnose based on a comprehensive understanding of a patient. In that way, a generalized patient representation can be obtained through pre-training. To achieve this goal in as robust a manner as possible, multi-task joint learning across different but related clinical outcomes is desired for composing the pre-training task.

We propose a novel representation learning architecture that combines supervised multi-task learning clinical phenotype prediction and fine-tuning on downstream tasks through transfer learning. Specifically, we pre-train the model from a large clinical note corpus with the supervised objectives of multiple clinical outcomes. The supervised objectives consist of different, but related high-prevalence diseases. We hypothesize this architecture is capable of learning a patient representation that is generalizable and universal, as multiple top high-prevalence phenotypes will cover the majority of patients.

To validate the robustness of the pre-trained representation, we apply it to be fine-tuned on low-prevalence disease phenotyping. Our primary motivation is to classify low-prevalence phenotyping with clinical notes. This task is generally challenging to achieve satisfactory performance because only a small number of positive samples are available. We hypothesize that through transfer learning, the pre-trained model will incorporate knowledge about patients and thus would improve the low-prevalence disease phenotyping as it is fine-tuned further. A generally-robust representation should largely favor "small data" tasks, where only a limited amount of samples are available to achieve optimal performance in prediction using its own data. Therefore, we consider the low-prevalence disease as a transfer learning target task to be further fine-tuned from the pre-trained multi-task source model to validate the generalizability of the source model.

Our experimental results suggest this generalized representation with multi-task pre-training consistently improves performance on low-prevalence phenotypes compared to single source task pre-training and traditional supervised learning. Importantly, we only need to pre-train once and use this generalized representation for numerous phenotypes. In practice, it is infeasible for NLP phenotyping models to always perform the array of phenotyping experiments that we demonstrate here. This is why we emphasize the stable and robust nature of the proposed method: it regularly outperforms the baselines, and in those cases where it does not, the performance delta is tolerably small. Thus, multi-task pre-training is a good starting point for future NLP phenotyping methods, especially for low-prevalence phenotypes.

## 2. RELATED WORK

Representation learning was first proposed to explore the motivations and techniques of learning good representations [4]. It was applied in the clinical domain and the techniques became more advanced and capable of modeling patient EHR data [1]. Some of the early attempts were proposed to generate computational representations based on patient diagnoses [5]. Apart from showing the feasibility of learning patient information, these studies also pointed out the challenges of handling EHR data, which is high-dimensional [6], sparse, heterogeneous [7], temporal [8], incomplete, less-interpretable [9,10], large-scale [11], and multimodal [12]. The motivation behind learning good patient representations is to encode meaningful information about the patient that can be used to predict various useful properties or outcomes including patient subtyping [13], mortality [14,15], readmission [16,17], computational phenotyping [18], and length of stay [11,12].

The earliest and most prominent work to learn patient representations from structured data in EHRs with deep learning models was DoctorAI [8], where a recurrent neural network was applied to learn patient representations from sequential clinical events. This work demonstrated the impact of transfer learning from large-scale clinical data to tasks with small amounts of data.

Numerous other studies have modeled patient data with supervised deep learning architectures including RETAIN [9], Dipole [19], HCNN [20], Patient2vec [21]. In addition to those supervised methods, DeepPatient [22] proposed an unsupervised learning approach, where the authors applied a stacked denoising autoencoder to learn patient representations from a longitudinal clinical data warehouse. Additionally, tensor decomposition is a deep learning approach used in unsupervised [23], supervised [24] and semi-supervised [25] manners. This method could learn patterns of interaction between events (e.g. beta blocker prescription => heart disease diagnosis) and was used to learn patient phenotype representations from EHR data [24].

Furthermore, with the success of learning effective word representations from natural language, many studies have been able to learn patient representations from unstructured data in EHRs. Similar to DeepPatient [22], Sushil et al. [26] explored unsupervised methods to learn patient representations from clinical notes and applied them as features to multiple prediction tasks. Dligach et al., [27] explored transfer learning where they pre-trained a CUI-based encoder from MIMIC-III [28] clinical notes and evaluated the encoder by extracting patient features to predict target phenotypes with a small number of patients. A more recent work by Kemp et al. [29] employed hierarchical modeling of clinical notes to build patient representations and additionally demonstrated the effect of pre-training. Steinberg et al. [30] proposed that language models are effective techniques for patient representation learning. Their experiments showed that adapting NLP techniques such as word2vec to patient representation schemes can increase the performance of clinical predictions, and the knowledge can be transferred from the entire patient cohort to the task-specific model. Overall, these studies support the hypothesis that learning a robust patient representation is feasible with resources directly from clinical notes and also highlight the potential of transfer learning of medical language.

Currently, studies using either multi-task learning or transfer learning to learn patient representation are still relatively novel. Although a few methods have applied multi-task learning to modeling clinical events with the assumption that certain patterns can be shared between related tasks [15,31], previous work has shown contradictory results [32]. Prior work on EHR transfer learning [8,27,33] has largely used a feature-based approach: extracting a patient vector from the intermediate layer of a deep learning model and using it as input to the downstream task. This extracting technique depends on task-customized models; thus, the performance improvement on the target task is incremental. Inspired by the idea of large source task pre-training and task-specific fine-tuning in open domain NLP [3,34], we proposed pre-training and fine-tuning to learn patient representations from medical language. To the best of our knowledge, this is one of the earliest attempts to build transferable and generalized patient representation from unstructured clinical notes, and to evaluate the representations on low-prevalence phenotypes.

## 3. METHODS

### 3.1 Proposed Method for Patient Representation Learning

We introduce the implementation of the transferable patient representation in this section. The overall architecture is shown in Figure 1. In general, the method consists of two steps: pre-training and fine-tuning. In pre-training, we apply multi-task learning in a supervised manner, sharing hidden information between multiple tasks. The model is jointly trained on different (but related) high-prevalence phenotypes. Our target task is to predict low-prevalence phenotypes.

Importantly, our work focuses on one of the most challenging clinical problems, low-prevalence phenotyping, as it is difficult to identify these phenotypes using their own limited data. So in fine-tuning, we consider the low-prevalence phenotype as the target task to be further fine-tuned. We select phenotypes that are relatively rare in MIMIC-III, specifically phenotypes with between 50 and 550 patients receiving the ICD-9 code. This results in 78 diseases that meet this low-prevalence threshold. During fine-tuning, the pre-trained model from multi-task supervised learning is restored and applied to the target task, and training is initialized with the pre-trained parameters. The model weights are continuously updated based on the target phenotype corresponding to its own labels. Each target task has its specific fully-connected layer before the final classification.

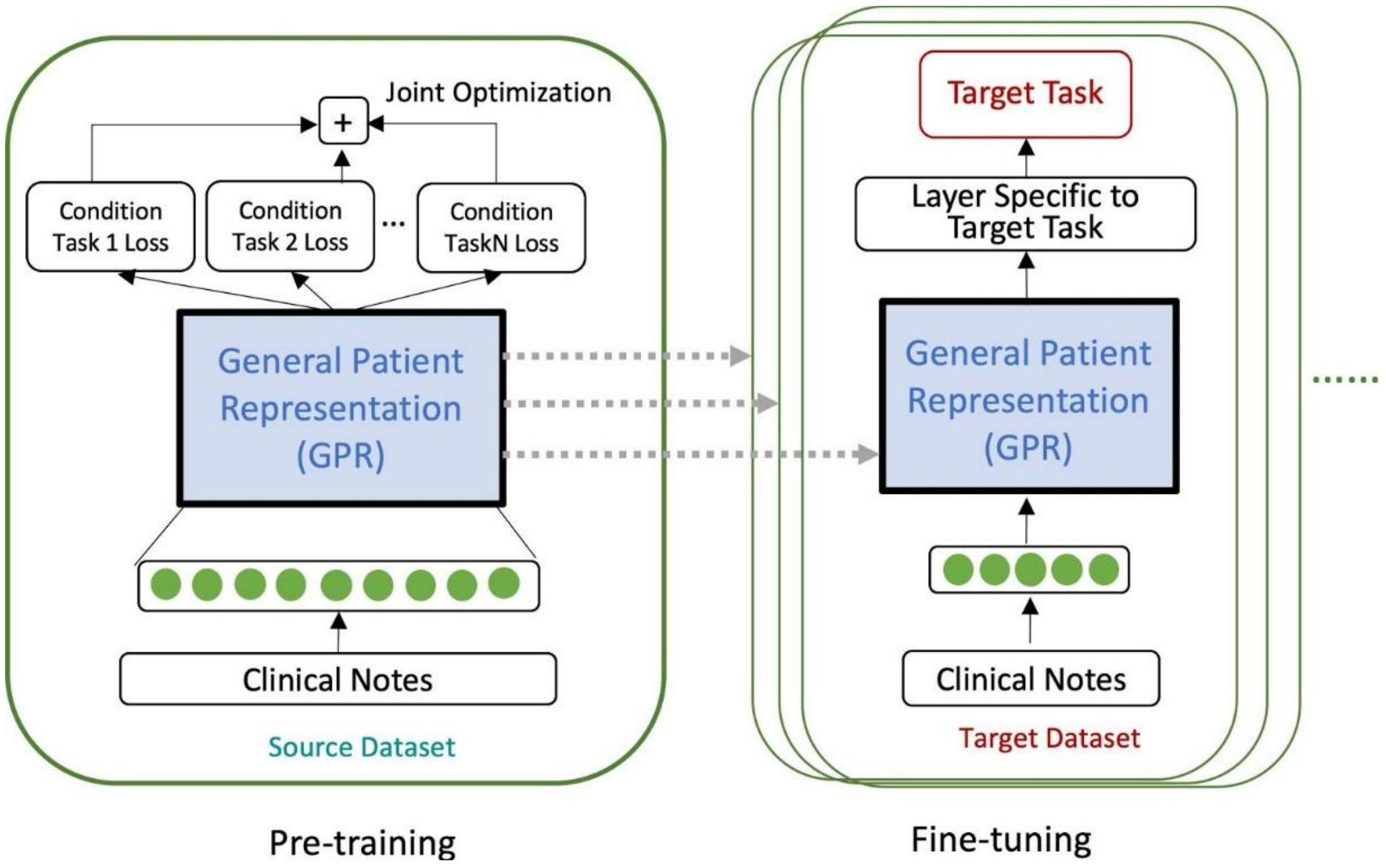

**Figure 1:** Overall pre-training and fine-tuning procedure for transferable patient representation learning. Each green box represents a learned model (one pre-trained model and multiple fine-tuned models).

The deep learning model for learning patient representations (i.e., General Patient Representation or GPR in Figure 1) is a hierarchical attention recurrent neural network (RNN) proposed in Si &

Roberts [33]. We consider this model because of its potential to deal with the hierarchical (i.e., word, sentence, document, patient) and chronological nature of the clinical note. The model is shown in Figure 2 and we will briefly describe the neural network. The network is trained progressively from word embeddings. At each level of hierarchy (sentence, document, and patient), we apply a bi-directional long-short term memory (LSTM) as the encoder and an attention mechanism. The output vector of each level is fed as the input of the next level, and the final output is directly applied to the outcome of either pre-training or fine-tuning. Also, at the patient level, we greedily combine notes within a time interval together, and separate those notes outside the interval, to capture the real temporal sequences between notes as a reflection of clinical reality. In reality, notes within a short time span often do not have a strong time order and they generally come in “bursts”. We choose a 1-hour interval in this work because the results of the 1-hour interval yielded the best performance in previous work [33]. As for the hyperparameters used in the model, we report results using the following: embedding size of 50; hidden and output size of LSTM: 100, and 100 respectively; the size of attention output: 200 at each hierarchy; and the total trainable parameters: 653,101. Note that we are comparing different methods rather than pursuing state-of-the-art performance, we choose this set of hyperparameters with preferences of both efficiency and effectiveness.

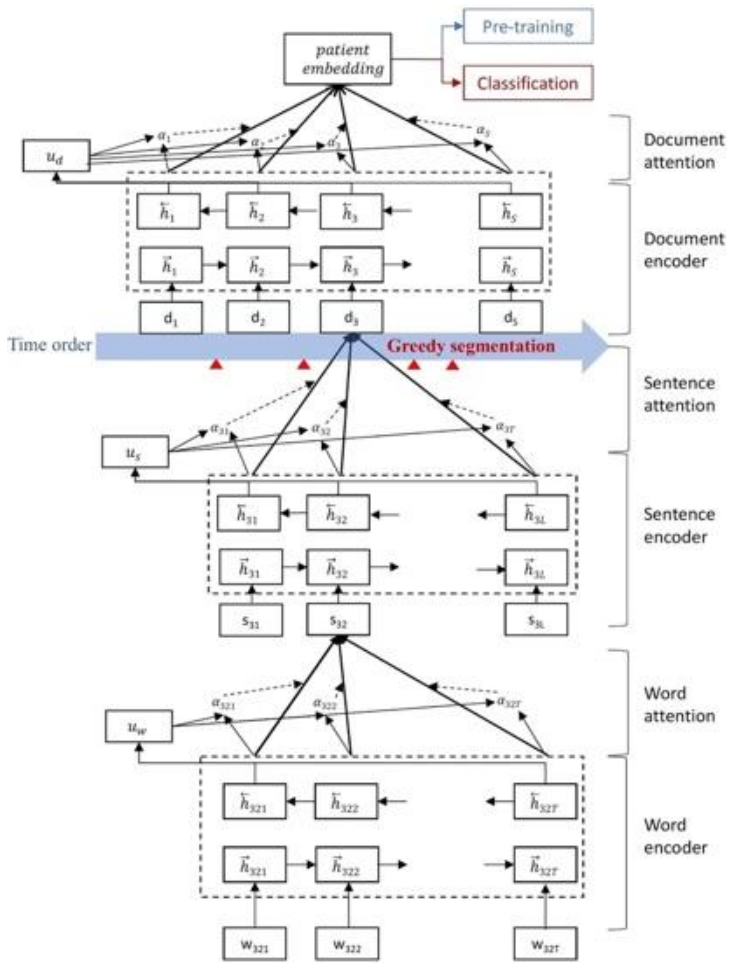

**Figure 2:** Hierarchical attention RNN architecture

### 3.2 High-Prevalence Phenotyping-guided Pre-training

To obtain a generalized patient representation, we construct a source task for pre-training to that attempts to be comprehensive by incorporating a wide variety of perspectives. We propose a multi-task joint training on multiple related high-prevalence phenotypes, so that patient information is more likely to be encoded in the pre-trained models. More specifically we looked into three organ systems: circulatory, respiratory, and genitourinary. These are three of the most frequent organ systems by cumulative ICD-9 code assignment. Within each organ system, we selected the top five phenotypes by patient frequency. We pre-train the model on the joint learning of these five high-prevalence phenotype tasks. The detailed information about phenotypes and the number of patients from MIMIC-III v1.4 [28] are shown in Table 1. This dataset is discussed in more detail below.

Table 1. Descriptive statistics of top five high-prevalence phenotypes in three organ systems.

| **Circulatory** | **Respiratory** | **Genitourinary** |
|---|---|---|

| Disease Name (ICD-9) | # patients | Disease Name (ICD-9) | # patients | Disease Name (ICD-9) | # patients |
|---|---|---|---|---|---|
| Unspecified essential hypertension (401.9) | 20703 | Acute respiratory failure (518.81) | 7497 | Acute kidney failure (584.9) | 9119 |
| Congestive heart failure (428.0) | 13111 | Pneumonia (486) | 4839 | Urinary tract infection (599.0) | 6555 |
| Atrial fibrillation (427.31) | 12891 | Chronic airway obstruction (496) | 4431 | Chronic kidney disease (585.9) | 3435 |
| Coronary atherosclerosis of native coronary artery (414.01) | 12429 | pleural effusion (511.9) | 2734 | Acute kidney failure with lesion of tubular necrosis (584.5) | 2287 |
| Hypertensive chronic kidney disease (403.90) | 3421 | Asthma (493.90) | 2195 | End stage renal disease (585.6) | 1926 |

We pre-train the deep learning model using clinical notes from MIMIC-III v1.4 [28], a public Intensive Critical Care database containing de-identified data from 46,520 patients, including 1.9 million clinical notes. As described in 3.1, with the greedy segmentation mechanism, for each patient, we combine clinical notes within 1 hour time intervals and separate clinical notes out of 1 hour spans into different units. At the word level, word embeddings obtained from our previous study [35] are applied to prepare input text features. The deep learning model adopted a three-level hierarchical attention RNN model (Figure 2) that progressively trains from input word embeddings to sentences, documents and finally towards the patient.

The outcomes for pre-training are to classify patients with high-prevalence phenotypes. Specifically, for each organ system, we pre-train a model on the joint learning of the top five phenotypes shown in Table 1. The loss/objective functions from all five tasks are jointly optimized to reach the minimum of the total loss. We use the ICD-9 code from the structured diagnosis table as the prediction label as a proxy for the phenotype. The labels for multi-task learning pre-training include the five highest-prevalence phenotypes. For example, if the patient has been diagnosed with disease A, B; not with disease C, D and E. The label for this patient should be A:1, B:1, C:0, D:0 and E:0. We can also consider the five-task pre-trained models as organ system-specific models, because each model specifically targets one organ system. Overall, we pre-trained three MTL models (i.e., Circulatory MTL model, Respiratory MTL model, and Genitourinary MTL model). For simplicity, we use the ICD-9 code from the structured diagnosis table as the prediction label as a proxy for the phenotype, but in future research, we will adopt more rigorous algorithms from medical resources like PheKB to identify the phenotype.

### 3.3 Fine-tuning on Rare Conditions

Fine-tuning is straightforward as we apply the pre-trained model to the target task, and fine-tune the parameters end-to-end. Specifically, the task-specific inputs including features from clinical notes and labels from structured diagnoses are fed into the pre-trained model. At the fine-tuning stage, the neural network is initialized with the weights from the pre-trained model. A fully-connected weight matrix is applied at the output layer for classification. The add-on part at the output layer for target task fine-tuning is similar to the fine-tuning for BERT [3], which transforms the output state from the neural network to a logit function that maps to a probability.

As stated, our main motivation for this study is to use a pre-training and fine-tuning approach to improve prediction for phenotypes that are relatively rare, as these would benefit the most from a robust transfer learning approach. To begin with, we select phenotypes that are relatively rare in MIMIC-III ($n$ = 50-550 patients) in three organ systems. In total, there are 78 diseases including 38 circulatory diseases, 23 respiratory diseases, and 17 genitourinary diseases that meet this low-prevalence threshold. We run experiments on all those diseases to avoid any cherry picking. The rare phenotype disease name, ICD-9 code, and the number of patients for each phenotype is shown in Table 2. Due to the extremely unbalanced distribution of positive samples for rare diseases, we assign a coefficient weight to the loss function to emphasize the positive targets over the negative to compensate for this bias. The formula for calculating the weight for each phenotype is defined as follows (we also report the weights reported in Table 2):

$$weight_pos = \frac{(1 \,/\, pos) \times total}{2}$$

pos: the number of patients who have positive labels;

total: the total number of patients (i.e., 31,360 in this experiment)

Table 2. Descriptive statistics of low-prevalence phenotypes in three organ systems.
(Weights: coefficient weights assigned with the loss to emphasize the positive samples. Most-related phenotype: one of the five high-prevalence phenotypes that is the most relevant to the target task.)

a. Circulatory

| Disease Name | ICD-9 | # cases | Weight | Most-related Phenotype |
|---|---|---|---|---|
| Acute systolic heart failure | 428.21 | 492 | 32 | 428.0 |
| Coronary atherosclerosis of autologous vein bypass graft | 414.02 | 474 | 33 | 414.01 |
| Other late effects of cerebrovascular disease | 438.89 | 465 | 34 | 401.9 |
| Benign essential hypertension | 401.1 | 454 | 35 | 401.9 |
| Late effects of cerebrovascular disease, hemiplegia affecting unspecified side | 438.20 | 437 | 36 | 403.90 |
| Acute diastolic heart failure | 428.31 | 432 | 36 | 401.9 |
| Systolic heart failure, unspecified | 428.20 | 416 | 38 | 428.0 |
| Subdural hemorrhage | 432.1 | 392 | 40 | 401.9 |
| Sinoatrial node dysfunction | 427.81 | 389 | 40 | 427.31 |
| Acute myocardial infarction of unspecified site, initial episode of care | 410.91 | 354 | 44 | 414.01 |
| Atherosclerosis of native arteries of the extremities with gangrene | 440.24 | 327 | 48 | 427.31 |
| Acute on chronic combined systolic and diastolic heart failure | 428.43 | 327 | 48 | 428.0 |
| Chronic total occlusion of coronary artery | 414.2 | 292 | 54 | 414.01 |
| Atherosclerosis of aorta | 440.0 | 283 | 55 | 414.01 |
| Sub-endocardial infarction, subsequent episode of care | 410.72 | 279 | 56 | 414.01 |
| Atherosclerosis of native arteries of the extremities with ulceration | 440.23 | 264 | 59 | 414.01 |
| Atherosclerosis of native arteries of the extremities with intermittent | 440.21 | 257 | 61 | 414.01 |
| claudication | | | | |
| Late effects of cerebrovascular disease, aphasia | 438.11 | 240 | 65 | 427.31 |
| Atherosclerosis of renal artery | 440.1 | 233 | 67 | 414.01 |
| Iatrogenic hypotension | 458.2 | 233 | 72 | 428.0 |
| Cerebral atherosclerosis | 437.0 | 196 | 80 | 401.9 |
| Hypertrophic cardiomyopathy | 425.1 | 183 | 86 | 401.9 |
| Chronic combined systolic and diastolic heart failure | 428.42 | 179 | 88 | 428.0 |
| Malignant essential hypertension | 401.0 | 172 | 91 | 401.9 |
| Paroxysmal supraventricular tachycardia | 427.0 | 161 | 97 | 427.31 |
| Acute myocardial infarction of anterolateral wall, initial episode of care | 410.01 | 142 | 110 | 414.01 |
| Hypertensive chronic kidney disease, benign, with chronic kidney disease stage I through stage IV, or unspecified | 403.10 | 122 | 129 | 403.90 |
| Combined systolic and diastolic heart failure, unspecified | 428.40 | 110 | 143 | 428.0 |
| Unspecified transient cerebral ischemia | 435.9 | 96 | 163 | 401.9 |
| Atherosclerosis of native arteries of the extremities with rest pain | 440.22 | 88 | 178 | 414.01 |
| Abdominal aneurysm, ruptured | 441.3 | 76 | 206 | 401.9 |
| Acute combined systolic and diastolic heart failure | 428.41 | 73 | 215 | 428.0 |
| Unspecified late effects of cerebrovascular disease | 438.9 | 69 | 227 | 401.9 |
| Unspecified cerebrovascular disease | 437.9 | 64 | 245 | 401.9 |
| Atherosclerosis of other specified arteries | 440.8 | 63 | 249 | 414.01 |
| Cerebral thrombosis with cerebral infarction | 434.01 | 60 | 261 | 401.9 |
| Secondary cardiomyopathy, unspecified | 425.9 | 53 | 296 | 428.0 |
| Acute myocardial infarction of unspecified site, subsequent episode of care | 410.92 | 53 | 296 | 414.01 |

b. Respiratory

| Disease Name | ICD-9 | # cases | Weight | Most-related Phenotype |
|---|---|---|---|---|
| Post-inflammatory pulmonary fibrosis | 515 | 544 | 29 | 486 |
| Pneumonia due to Pseudomonas | 482.1 | 430 | 36 | 486 |
| Chronic respiratory failure | 518.83 | 331 | 47 | 518.81 |
| Acute edema of lung, unspecified | 518.4 | 305 | 51 | 511.9 |
| Chronic obstructive asthma with (acute) exacerbation | 493.22 | 299 | 52 | 496 |
| Pneumonia due to other gram-negative bacteria | 482.83 | 264 | 59 | 486 |
| Bacterial pneumonia, unspecified | 482.9 | 227 | 69 | 486 |
| Pneumonia due to Klebsiella pneumoniae | 482.0 | 226 | 69 | 486 |
| Pneumococcal pneumonia [Streptococcus pneumoniae pneumonia] | 481 | 194 | 81 | 486 |
| Bronchiectasis without acute exacerbation | 494.0 | 191 | 82 | 486 |
| Empyema without mention of fistula | 510.9 | 190 | 83 | 511.9 |
| Methicillin resistant pneumonia due to Staphylococcus aureus | 482.42 | 162 | 97 | 518.81 |
| Pulmonary congestion and hypostasis | 514 | 155 | 101 | 511.9 |
| Unspecified sinusitis (chronic) | 473.9 | 149 | 105 | 493.90 |
| Edema of larynx | 478.6 | 145 | 108 | 518.81 |
| Malignant pleural effusion | 511.81 | 132 | 119 | 511.9 |
| Acute bronchitis | 466.0 | 126 | 124 | 518.81 |
| Asbestosis | 501 | 116 | 135 | 496 |
| Acute upper respiratory infections of unspecified site | 465.9 | 96 | 163 | 493.90 |
| Abscess of lung | 513.0 | 86 | 182 | 486 |
| Unilateral paralysis of vocal cords or larynx, partial | 478.31 | 74 | 212 | 518.81 |
| Empyema with fistula | 510.0 | 72 | 218 | 511.9 |
| Stenosis of larynx | 478.74 | 61 | 257 | 518.81 |

c. Genitourinary

| **Disease Name** | **ICD-9** | **# cases** | **Weight** | **Most-related Phenotype** |
|---|---|---|---|---|
| Hematuria | 599.7 | 509 | 31 | 599.0 |
| Hydronephrosis | 591 | 413 | 38 | 584.9 |
| Chronic kidney disease, Stage IV (severe) | 585.4 | 334 | 47 | 585.6 |
| Hypertrophy (benign) of prostate with urinary obstruction and other lower urinary tract symptoms (LUTS) | 600.01 | 314 | 50 | 599.0 |
| Neurogenic bladder NOS | 596.54 | 225 | 70 | 599.0 |
| Hematuria, unspecified | 599.70 | 216 | 73 | 599.0 |
| Calculus of kidney | 592.0 | 206 | 76 | 599.0 |
| Gross hematuria | 599.71 | 181 | 87 | 599.0 |
| Secondary hyperparathyroidism (of renal origin) | 588.81 | 169 | 93 | 585.9 |
| Acute pyelonephritis without lesion of renal medullary necrosis | 590.10 | 132 | 119 | 599.0 |
| Calculus of ureter | 592.1 | 111 | 141 | 599.0 |
| Cyst of kidney, acquired | 593.2 | 107 | 147 | 585.9 |
| Hypertrophy (benign) of prostate | 600.0 | 92 | 170 | 599.0 |
| Pyelonephritis, unspecified | 590.80 | 84 | 187 | 599.0 |
| Nephritis and nephropathy, not specified as acute or chronic, with unspecified pathological lesion in kidney | 583.9 | 84 | 187 | 584.9 |
| Vascular disorders of kidney | 593.81 | 83 | 189 | 584.5 |
| Chronic glomerulonephritis in diseases classified elsewhere | 582.81 | 67 | 234 | 585.9 |

### 3.4 Baseline Approaches

To evaluate the proposed method, we compare the predictive performance with the following three baseline methods:

**[Baseline 1] Single-task pre-training with top one high-prevalence phenotype:** The pre-training source task consists of only one of the five high-prevalence phenotypes with the highest number of patients. More specifically, the pre-training task for circulatory, respiratory, and genitourinary corresponds to *Unspecified essential hypertension (ICD-9: 401.9)*, *Acute respiratory failure (ICD-9: 518.81)*, and *Acute kidney failure (ICD-9: 584.9)*, respectively.

**[Baseline 2] Single-task pre-training with the most relevant high-prevalence phenotype:** The pre-training source task consists of only one of the five high-prevalence phenotypes that is the most related to the target task. The selection of choosing the most related source phenotype was made by a practicing clinician in internal medicine (EVB) based on clinical knowledge. The selection is prior to training the pre-trained model. The target phenotype with its most related high-prevalence phenotype (out of the five) is shown in the column of "Most-related Phenotype" in Table 2.

**[Baseline 3] Target only:** This is a traditional deep learning method and the model is trained solely based on the specific target task for each low-prevalence phenotype.

For simplicity, in the following sections, we name the proposed method, multi-task learning as "MTL"; baseline 1 method, single-task learning with the top one high-prevalence phenotype as "STL-highest"; baseline 2 method, single-task learning with the most clinically-related phenotype as "STL-related"; and baseline 3 method, traditional machine learning as "Target".

### 3.5 Data and Experimental Details

We conduct the experiments using clinical notes from MIMIC-III v1.4 [28], a collection of 2 million free-text notes. We only keep adults (age $\geq$ 18), and delete notes with the "ISERROR" label. For each note, basic tokenization with regular expression and sentence segmentation with spaCy are performed. We apply word2vec embeddings (dimension of 50) trained in our previous study [35] to represent tokens and to prepare the input text features. For each patient, notes within 1-hour are combined together as one note and notes outside the 1-hour span are splitted into different segments. The structured diagnosis tables consisting of ICD-9 codes are used to construct the prediction label for patients. This results in 31,360 patients to use in the experiments below. We utilize ICD-9 codes as the prediction label and a proxy for the phenotype. For a given ICD-9 code, if the patient has the ICD-9 code on record, then this patient is considered a positive case for that ICD-9 phenotype. All other patients without being labeled with the ICD-9 code in structured diagnosis tables are negative cases.

For each patient and each low-prevalence phenotype, the input features for pre-training and fine-tuning are the embeddings derived from the patient's clinical notes. But the prediction labels in pre-training depend on the pre-training method. The labels for "MTL" consist of the five highest-prevalence phenotypes in each organ system. The labels for "STL-highest" come from the highest-prevalence phenotype for each organ system (listed in section 3.4, Baseline 1). The labels for “STL-related” are constructed from Table 2 column “Most-related Phenotype”. There is no pre-training for “Target” and the labels for “Target” are solely whether the patient has been diagnosed with each low-prevalence disease. That was also the label for fine-tuning across all pre-training methods. The predictions for pre-training and fine-tuning are altogether patient-level.

We set aside non-overlapping 80%, 10%, 10% samples at random from the entire dataset as training, validation, and test sets, respectively. The validation set is used to evaluate the trained model and prevent the model from overfitting on training sets. We applied the early stopping technique on validation sets to prevent overfitting. More specifically, the validation loss is calculated at each epoch. Early stopping is triggered when there is no improvement in the validation loss (i.e., the loss is increasing) for three consecutive epochs. To avoid any possible data leakage, we reserve the test set and only use it to report the performance. In other words, the pre-training or fine-tuning process does not have any chance to learn test data. Both the pre-training and fine-tuning task use the hierarchical attention recurrent neural network to predict and build the general patient representation. Networks were trained using Adam optimization (batch size of 32, learning rate of 0.001). We use a sigmoid loss function and the positive weights from Table 2. We report the area under the ROC curve (AUC) on the test set for all four methods (one proposed and three baselines) and for all 78 diseases (from three organ systems), which results in 312 AUC scores. We note that small test sets for some of the phenotypes is a limitation of this work, leading to greater variance in the results. However, the use of AUC instead of F1 measure reduces the variance in some degree as AUC is a more stable metric.

To compare different pre-training methods, we also report:

1. the AUC scores across all low prevalence phenotypes (Table 4);
2. box plots of the performance distribution in different organ systems (Figure 3);
3. the number of phenotypes that the proposed method outperformed the baselines (Table 5);
4. for each method, the number of phenotypes for which the method achieves the best performance (Table 6);
5. when the method does not achieve the best result, the number of phenotypes that are still within 90% of the best performance (Table 6);
6. the average mean squared error on test set for each method (Table 7).

To better understand the time and computational resources dedicated for future reproducibility of the experiments, we investigate the time for pre-training with different configuration settings (i.e., MTL, STL), for fine-tuning, and for target-only training. In general, 5-task pre-training (i.e., proposed MTL) took around 10 minutes for each epoch. Single-task pre-training took around 9 minutes every epoch. Given a defined phenotype, fine-tuning initialized with the 5-task model took around 3 minutes every epoch, and fine-tuning initialized with the STL model took around 5 minutes per epoch. Target-only method took around 9 minutes for each epoch, similar to MTL/STL, because the model is exactly the same in pre-training. In total, MTL pre-training took around 3 hours, and STL pre-training took around 2.5 hours.

## 4. RESULTS

### 4.1 Pre-training on high-prevalence phenotypes

We pre-trained three organ system-specific models using MTL of the top five highest-prevalence phenotypes within the same organ system. Specifically, we used three five-task pre-trained models (i.e., 3 MTL models including Circulatory MTL model, Respiratory MTL model, Genitourinary MTL model). We also pre-trained single-task learning (STL) models for each phenotype (15 STL models). Performances (AUCs) of MTL and STL for each phenotype are reported in Table 3. Here the MTL column represents that the pre-trained MTL model was tested on the test set for each high-prevalence phenotype. The STL column reports the performance on the test set using pre-trained STL models. Specifically, for each organ system, the score in the STL column on each row was obtained from one STL model, while the five scores in the MTL column were coming from one MTL model. Although there was only one model in MTL pre-training, as five phenotypes were used to optimize in the training; thus there would be five AUC scores reported using the test set. Although we focused on low-prevalence phenotypes, we noted that MTL did not harm the performance on high-prevalence phenotypes and achieved a well-matched performance among all tasks. Notably, there was a distinct improvement from STL to MTL for the majority of phenotypes, with the biggest improvement (0.2715) from STL to MTL on Asthma.

Table 3. Performance in AUC scores for high-prevalence phenotypes with different methods of MTL and STL. Within the same organ system, the performance of each disease in the MTL column comes from one MTL model, while the performance of each disease in the STL column comes from the individual STL model.

| Circulatory | MTL | STL | Respiratory | MTL | STL | Genitourinary | MTL | STL |
|---|---|---|---|---|---|---|---|---|
| Essential hypertension | 0.8041 | 0.8041 | Acute respiratory failure | 0.9107 | **0.9112** | Acute kidney failure | 0.8469 | **0.8519** |
| Congestive heart failure | **0.9183** | 0.9145 | Pneumonia | 0.8542 | **0.8603** | Urinary tract infection | 0.7423 | **0.7468** |
| Atrial fibrillation | **0.9408** | 0.9336 | Chronic airway obstruction | **0.8378** | 0.8048 | Chronic kidney disease | **0.8664** | 0.8550 |
| Coronary atherosclerosis of native coronary artery | **0.9517** | 0.9503 | Pleural effusion | 0.8539 | **0.8560** | Acute kidney failure with lesion of tubular necrosis | **0.9105** | 0.8882 |
| Hypertensive chronic kidney disease | **0.8768** | 0.8731 | Asthma | **0.8449** | 0.5734 | End stage renal | **0.9752** | 0.9412 |

## 4.2 Transfer learning on low-prevalence phenotypes

### 4.2.1 The effectiveness of pre-training

We further investigated the effectiveness of pre-training and fine-tuning on the low-prevalence phenotypes. To get an overall sense of the effectiveness of the proposed method and three baselines, we plotted the distribution of AUC scores for four methods among three organ systems in Figure 3. The AUC performances of three organ systems for all 78 phenotypes are reported in Table 4.

In Figure 3, each box plot represented the performance for each organ system. The box plots showed the median, first and third quartile, minimum, and maximum of scores for different methods. We noticed that among four methods, the MTL performances were the most compact and Target performances were the most variable. We also observed a trend for each organ system that the median of MTL was higher than that of the other three methods.

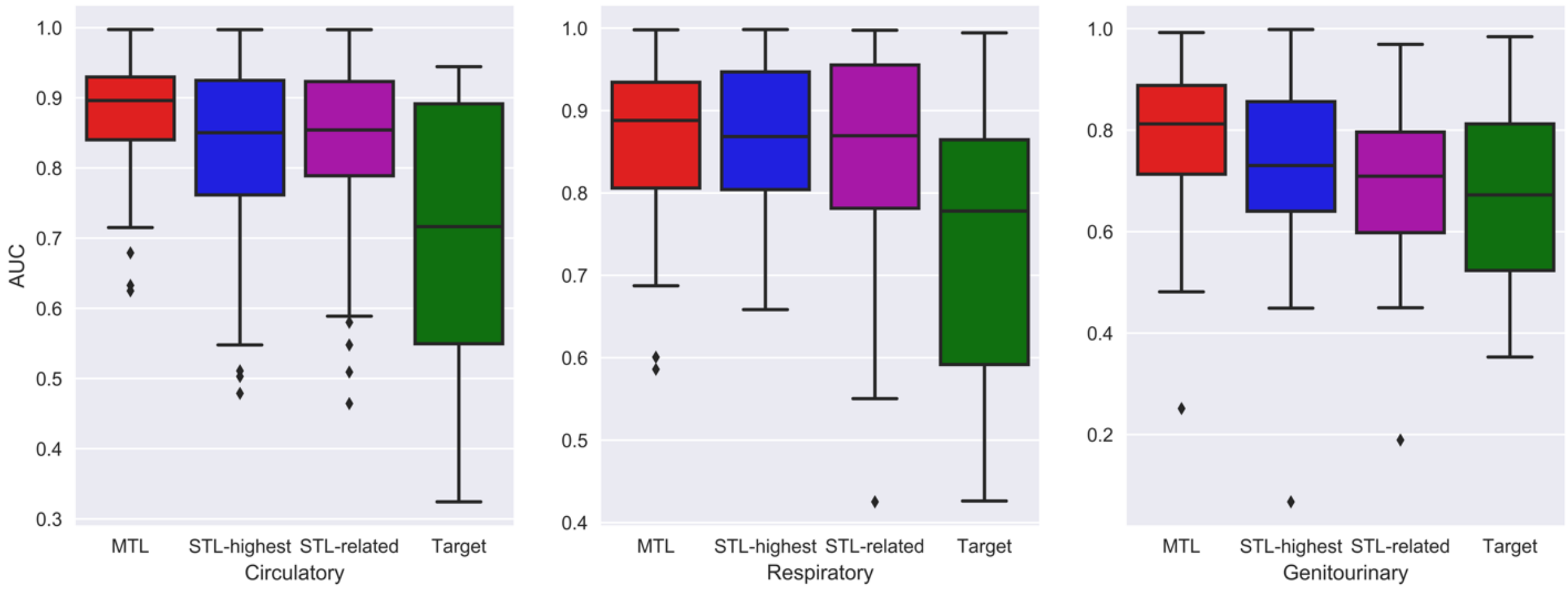

**Figure 3:** Box plots of AUC performance across different methods among three organ systems.

In Table 4, for 86% of diseases (67 out of 78), at least one of the three pre-training methods (*MTL*, *STL-highest*, *STL-related*) worked better than straightforward *Target* prediction. Specifically, *MTL* performed better than *Target* in 28 circulatory phenotypes, 18 respiratory phenotypes, and 11 genitourinary phenotypes. *STL-highest* is higher than *Target* in 26 circulatory phenotypes, 20 respiratory phenotypes, and 10 genitourinary phenotypes. *STL-related* performed better than *Target* in 27 circulatory phenotypes, 18 respiratory phenotypes, and 9 genitourinary phenotypes. Specifically, we saw the biggest jumps from three pre-training methods to solely *Target* are 0.5946 in the circulatory system (Disease Name: Atherosclerosis of other specified arteries, ICD-9: 440.8), 0.4860 in the respiratory system (Disease Name: Chronic obstructive asthma with exacerbation, ICD-9: 493.22), 0.5159 in the genitourinary system (Disease Name: Chronic kidney disease, Stage IV severe, ICD-9: 585.4).

**Table 4.** AUC of different methods on target phenotyping tasks in organ systems (AUC in bold means the best across four methods).

A. Circulatory

| Disease Name | MTL | STL-highest | STL-related | Target |
|---|---|---|---|---|
| Acute systolic heart failure | 0.9256 | 0.9238 | 0.9124 | **0.9279** |
| Coronary atherosclerosis of autologous vein bypass graft | 0.9627 | **0.9662** | 0.9403 | 0.9295 |
| Other late effects of cerebrovascular disease | **0.8595** | 0.8301 | | 0.5741 |
| Benign essential hypertension | 0.8542 | **0.8850** | | 0.6578 |
| Late effects of cerebrovascular disease, hemiplegia affecting unspecified side | **0.9533** | 0.9297 | 0.9057 | 0.4138 |
| Acute diastolic heart failure | 0.8801 | 0.8793 | | **0.9086** |
| Systolic heart failure, unspecified | 0.8248 | **0.8668** | 0.5946 | 0.7846 |
| Subdural hemorrhage | 0.9036 | **0.9640** | | 0.9063 |
| Sinoatrial node dysfunction | 0.9310 | 0.9251 | **0.9359** | 0.9209 |
| Acute myocardial infarction of unspecified site, initial episode of care | **0.8400** | 0.8151 | 0.7982 | 0.7581 |
| Atherosclerosis of native arteries of the extremities with gangrene | 0.9448 | 0.9305 | **0.9557** | 0.9241 |
| Acute on chronic combined systolic and diastolic heart failure | 0.8448 | 0.8173 | 0.8330 | **0.8594** |
| Chronic total occlusion of coronary artery | 0.9509 | 0.9405 | **0.9532** | 0.9444 |
| Atherosclerosis of aorta | 0.8158 | 0.8345 | 0.8440 | **0.8709** |
| Sub-endocardial infarction, subsequent episode of care | **0.9396** | 0.8986 | 0.9043 | 0.4282 |
| Atherosclerosis of native arteries of the extremities with ulceration | 0.8312 | 0.8341 | **0.9213** | 0.6387 |
| Atherosclerosis of native arteries of the extremities with intermittent claudication | 0.8295 | **0.9059** | 0.8659 | 0.8893 |
| Late effects of cerebrovascular disease, aphasia | **0.9590** | 0.7527 | 0.6746 | 0.5184 |
| Atherosclerosis of renal artery | **0.9014** | 0.8544 | 0.7915 | 0.4861 |
| Iatrogenic hypotension | **0.9223** | 0.8896 | 0.8526 | 0.6864 |
| Cerebral atherosclerosis | **0.7836** | 0.7524 | | 0.5232 |
| Hypertrophic cardiomyopathy | **0.9951** | 0.9378 | | 0.6124 |
| Chronic combined systolic and diastolic heart failure | **0.9043** | 0.5110 | 0.8916 | 0.8922 |
| Malignant essential hypertension | **0.8528** | 0.8480 | | 0.6048 |
| Paroxysmal supraventricular tachycardia | **0.7152** | 0.6640 | 0.5093 | 0.7146 |
| Acute myocardial infarction of anterolateral wall, initial episode of care | 0.9124 | 0.5029 | **0.9240** | 0.9172 |
| Hypertensive chronic kidney disease, benign, with chronic kidney disease stage I through stage IV, or unspecified | **0.8408** | 0.7920 | 0.4643 | 0.3244 |
| Combined systolic and diastolic heart failure, unspecified | 0.8715 | 0.6269 | 0.8559 | **0.8998** |
| Unspecified transient cerebral ischemia | 0.6328 | 0.5479 | | **0.6945** |
| Atherosclerosis of native arteries of the extremities with rest pain | **0.9041** | 0.7885 | 0.5801 | 0.5415 |
| Abdominal aneurysm, ruptured | **0.9974** | 0.9972 | | 0.5881 |
| Acute combined systolic and diastolic heart failure | 0.9204 | 0.8526 | **0.9426** | 0.7184 |
| Unspecified late effects of cerebrovascular disease | 0.6250 | **0.8097** | | 0.7283 |
| Unspecified cerebrovascular disease | **0.8909** | 0.5888 | | 0.8324 |
| Atherosclerosis of other specified arteries | 0.9216 | **0.9519** | 0.7880 | 0.3573 |
| Cerebral thrombosis with cerebral infarction | **0.9381** | 0.9351 | | 0.7420 |
| Secondary cardiomyopathy, unspecified | **0.6790** | 0.4789 | 0.6423 | 0.5308 |
| Acute myocardial infarction of unspecified site, subsequent episode of care | **0.8799** | 0.6348 | 0.8387 | 0.4825 |

B. Respiratory

| **Disease Name** | **MTL** | **STL-highest** | **STL-related** | **Target** |
|---|---|---|---|---|
| Post-inflammatory pulmonary fibrosis | **0.9194** | 0.8437 | 0.8694 | 0.8487 |
| Pneumonia due to Pseudomonas | **0.9381** | 0.9307 | 0.9267 | 0.9186 |
| Chronic respiratory failure | **0.8879** | 0.7741 | | 0.7496 |
| Acute edema of lung, unspecified | 0.8358 | 0.8393 | 0.8332 | **0.8662** |
| Chronic obstructive asthma with (acute) exacerbation | 0.9866 | **0.9880** | 0.9743 | 0.5019 |
| Pneumonia due to other gram-negative bacteria | **0.9306** | 0.9074 | 0.9271 | 0.9198 |
| Bacterial pneumonia, unspecified | **0.8862** | 0.8831 | 0.8773 | 0.8629 |
| Pneumonia due to Klebsiella pneumoniae | **0.8858** | 0.8685 | 0.8855 | 0.8179 |
| Pneumococcal pneumonia [Streptococcus pneumoniae pneumonia] | 0.8029 | **0.8097** | 0.7887 | 0.7833 |
| Bronchiectasis without acute exacerbation | 0.6873 | **0.7980** | 0.6988 | 0.7016 |
| Empyema without mention of fistula | 0.9858 | **0.9959** | 0.9888 | 0.9873 |
| Methicillin resistant pneumonia due to Staphylococcus aureus | 0.9291 | **0.9600** | | 0.6498 |
| Pulmonary congestion and hypostasis | 0.6007 | **0.7987** | 0.5504 | 0.6108 |
| Unspecified sinusitis (chronic) | 0.5860 | **0.6584** | 0.6340 | 0.6365 |
| Edema of larynx | 0.8221 | **0.8302** | | 0.5535 |
| Malignant pleural effusion | **0.9767** | 0.9266 | 0.9842 | 0.8537 |
| Acute bronchitis | 0.7482 | **0.8139** | | 0.4523 |
| Asbestosis | **0.7506** | 0.7375 | 0.4252 | 0.4262 |
| Acute upper respiratory infections of unspecified site | 0.9289 | **0.9380** | 0.8356 | 0.5727 |
| Abscess of lung | 0.8882 | **0.9557** | 0.9505 | 0.7781 |
| Unilateral paralysis of vocal cords or larynx, partial | **0.8090** | 0.6864 | | 0.4811 |
| Empyema with fistula | 0.9981 | **0.9984** | 0.9879 | 0.9875 |
| Stenosis of larynx | **0.9979** | 0.9976 | | 0.9943 |

C. Genitourinary

| Disease Name | MTL | STL-highest | STL-related | Target |
|---|---|---|---|---|
| Hematuria | **0.8609** | 0.8168 | 0.7640 | 0.8124 |
| Hydronephrosis | 0.9513 | **0.9690** | | 0.9690 |
| Chronic kidney disease, Stage IV (severe) | 0.8124 | **0.8687** | 0.5444 | 0.3528 |
| Hypertrophy (benign) of prostate with urinary obstruction and other lower urinary tract symptoms (LUTS) | 0.7134 | 0.7252 | **0.7542** | 0.6236 |
| Neurogenic bladder NOS | **0.6845** | 0.5690 | 0.6133 | 0.4658 |
| Hematuria, unspecified | **0.7362** | 0.6404 | 0.7012 | 0.6723 |
| Calculus of kidney | **0.8299** | 0.6291 | 0.6624 | 0.7257 |
| Gross hematuria | 0.8590 | 0.8347 | **0.8857** | 0.8100 |
| Secondary hyperparathyroidism (of renal origin) | 0.7201 | **0.7595** | 0.5630 | 0.5462 |
| Acute pyelonephritis without lesion of renal medullary necrosis | 0.9534 | **0.9710** | 0.9528 | 0.9218 |
| Calculus of ureter | 0.9925 | **0.9984** | 0.5979 | 0.4971 |
| Cyst of kidney, acquired | 0.4814 | **0.7061** | 0.4498 | 0.5582 |
| Hypertrophy (benign) of prostate | 0.8882 | 0.4491 | 0.9069 | **0.9320** |
| Pyelonephritis, unspecified | **0.9189** | 0.8562 | 0.7573 | 0.5235 |
| Nephritis and nephropathy, not specified as acute or chronic, with unspecified pathological lesion in kidney | 0.7017 | 0.7093 | | **0.7848** |
| Vascular disorders of kidney | 0.7180 | 0.7303 | 0.7963 | **0.9842** |
| Chronic glomerulonephritis in diseases classified elsewhere | 0.2514 | 0.0676 | 0.1894 | **0.5203** |

### 4.2.2 The effectiveness of MTL

The number of phenotypes in which the proposed method outperformed the baseline methods are shown in Table 5. Totally, the number of phenotypes that *MTL* outperforming the other baseline are always higher than that in the opposite way (i.e., the other baseline outperformed MTL). Specifically, 47 phenotypes (60%) have higher AUC scores with *MTL* over *STL-highest*, 53 phenotypes (68%) have higher AUC scores with *MTL* over *STL-related*, and 57 phenotypes (73%) have higher AUC scores with *MTL* over *Target*. We can also refer to Table 4 and find out that the biggest improvements of *MTL* over the maximum of the other three methods were 0.2063 in the circulatory system (ICD-9: 438.11, *MTL*: 0.959, *STL-highest*: 0.7527), 0.1226 in the respiratory system (ICD-9: 478.31, *MTL*: 0.809, *STL-highest/related*: 0.6864), and 0.1042 in the genitourinary system (ICD-9: 592.0, *MTL*: 0.8299, *Target*: 0.7257).

**Table 5.** Comparisons of MTL with other Baselines in Three Organ Systems

| # phenotypes | *MTL & STL-highest* | | *MTL & STL-related* | | *MTL & Target* | |
|---|---|---|---|---|---|---|
| | *MTL > STL-highest* | *STL-highest >MTL* | *MTL > STL-related* | *STL-related >MTL* | *MTL >Target* | *Target >MTL* |
| Circulatory | 29 | 9 | 27 | 11 | 28 | 10 |
| Respiratory | 10 | 13 | 15 | 8 | 18 | 5 |
| Genitourinary | 8 | 9 | 11 | 6 | 11 | 6 |
| Total | **47** | 31 | **53** | 25 | **57** | 21 |

*MTL* pre-training improved performance consistently and often substantially over single-task pre-training (*STL-highest* and *STL-related*). In Table 6 we summarized that, among 78 phenotypes, 33 phenotypes obtained the best performance with *MTL* (circulatory: 19; respiratory: 9; genitourinary: 5); 25 phenotypes obtained the best performance with *STL-highest* (circulatory: 7; respiratory: 12; genitourinary: 6); 9 phenotypes obtained the best performance with *STL-related* (circulatory: 6; respiratory: 1; genitourinary: 2); and 11 phenotypes obtained the best performance with *Target* (circulatory: n=6; respiratory: n=1; genitourinary: n=4). These 11 phenotypes are shown in Appendix Table 1. As shown in Supplemental Table 1, for the majority of phenotypes where target-only training outperformed the other methods, we still observed that they achieved comparative results with MTL. For the remaining, our assumption is that for some phenotypes such as *Chronic glomerulonephritis in diseases classified elsewhere*, because the sample size is too limited, even the recipe of pre-training and fine-tuning is still not helpful enough to improve the prediction. More importantly, the sample size is the key factor.

When the approach did not perform the best, if it was within 90% of the best performance, we considered this *tolerable* because the performance differences were not likely to be practically significant. Thus, we count the number of tolerable cases for four different methods. For the circulatory, respiratory, and genitourinary systems, the number of phenotypes that *MTL* has the best performance was 19, 9, 5, respectively. For phenotypes where *MTL* not have the best performance, the number of phenotypes where *MTL* performance was tolerable (within 90% of the best performance) was 18 (out of 19), 11 (out of 14), 8 (out of 12), respectively. As shown in Table 6, among four methods, *MTL* obtained the most number of best performance, and *MTL* also achieved the most number of tolerable cases.

**Table 6.** Tolerable Performance Analysis
[The percent is the second column was divided by the total number of phenotypes, which is 78.
The percent in the third column was divided by the number of phenotypes not performing the best. That number is 45, 53, 69, 67 for *MTL*, *STL-highest*, *STL-related*, *Target*, respectively.]

| | Best performance count (percent) | Within 90% of the best performance count (percent) |
|---|---|---|
| MTL | 33 (42.31%) | 37 (82.22%) |
| STL-highest | 25 (32.05%) | 36 (67.92%) |
| STL-related | 9 (11.54%) | 44 (63.77%) |
| Target | 11 (14.1%) | 27 (40.30%) |

To quantitatively measure the effectiveness of *MTL*, we calculated the average mean squared error on the test set (Table 7). The average mean squared error is inversely proportional to the performance. What stands out in Table 7 is that for the three organ systems, the average mean squared error of *MTL* was smallest compared to the other three methods.

**Table 7.** Average Mean Squared Error on Test set across Three Organ Systems

| | *MTL* | *STL-high* | *STL-related* | *Target* |
|---|---|---|---|---|
| Circulatory | 0.0368 | 0.0488 | 0.0477 | 0.0490 |
| Respiratory | 0.0388 | 0.0461 | 0.0397 | 0.0422 |
| Genitourinary | 0.0384 | 0.0431 | 0.0447 | 0.0500 |

Overall, in the cases where *MTL* performed worse than one of these baselines, its performance was still usually close, while in cases where it outperforms a baseline its performance was often much better. The above behaviors justify our hypothesis that *MTL* pre-training is generally robust in terms of the prediction performance on low-prevalence phenotypes.

#### 4.2.3 Efficiency of MTL

We assume the proposed method can also learn faster while improving the performance for some phenotypes in target tasks. Thus, we plotted the performance on the validation set during training to show how many epochs were required for convergence (stopping) for each model. We assume training would be more efficient if the model already contained existing knowledge from other resources and the hyperparameters from the pre-trained model would be reflected on the target task fine-tuning. In other words, the model would take fewer epochs to reach optimal performance if using the pre-trained model.

We randomly selected two phenotypes and plot performance (AUC) on validation sets to show the learning efficiency of the four methods across training epochs. As shown in Figure 4(a), the phenotype was related to Acute myocardial infarction (ICD-9: 410.91), *MTL* pre-training took 6 epochs to achieve the best performance, or the lowest loss. *STL-highest* pre-training also took 6, and *STL-related* took 8 epochs. Target task without any pre-training required 13 epochs and does not reach best performance (AUC scores referred to Table 4 A).

Figure 4(b) is another example from the genitourinary system, the phenotype is related to Hematuria (ICD-9: 599.70), *MTL* pre-training took 4 epochs to achieve the best performance.

*STL-highest* and *STL-related* pre-training took 8 and 7 epochs, respectively. *Target* task without any pre-training took 9 epochs (AUC scores referred to Table 4 C).

a. Disease Name: Acute myocardial infarction of unspecified site, initial episode of care (ICD-9: 410.91)

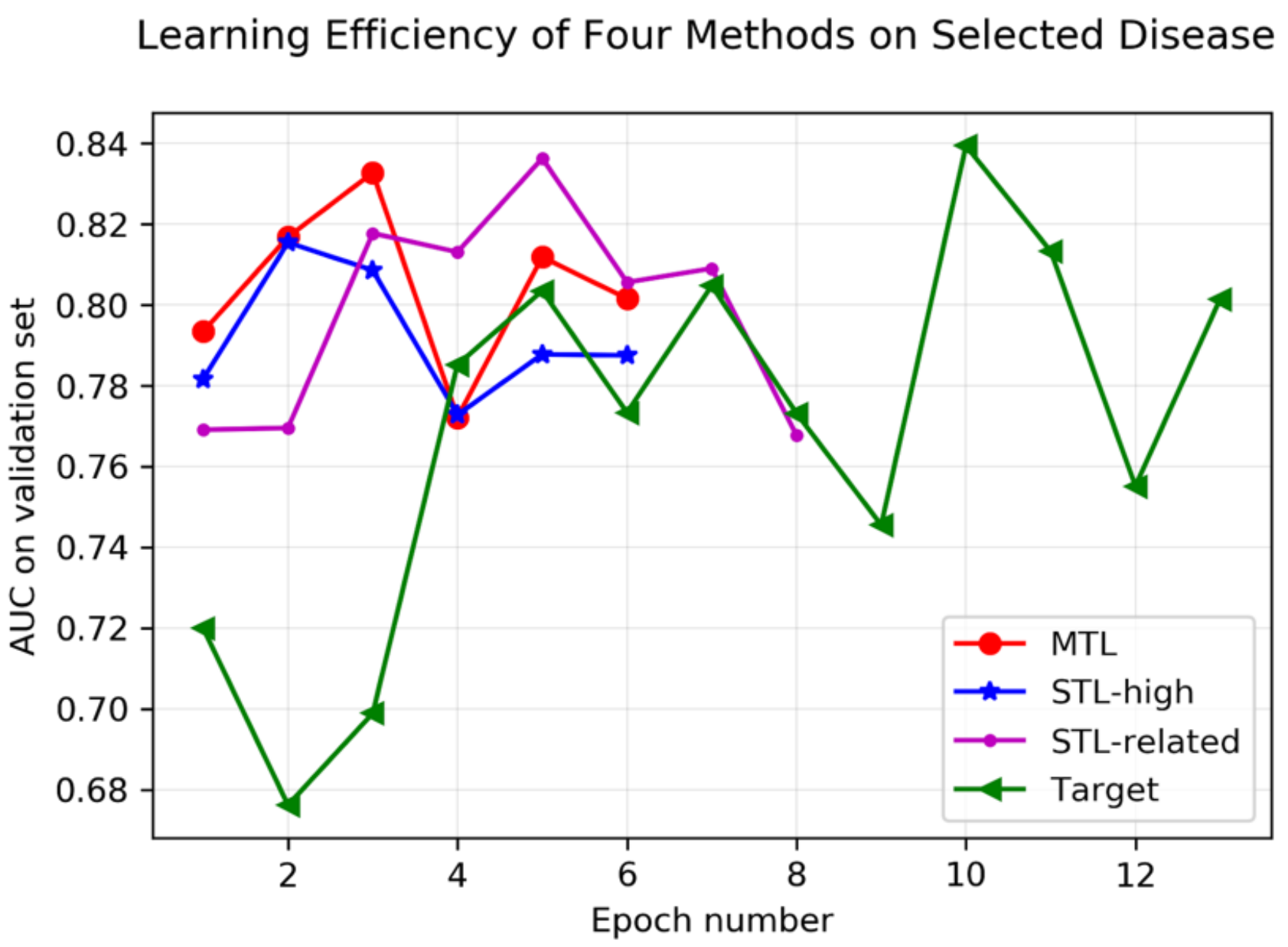

b. Disease Name: Hematuria, unspecified (ICD-9: 599.70)

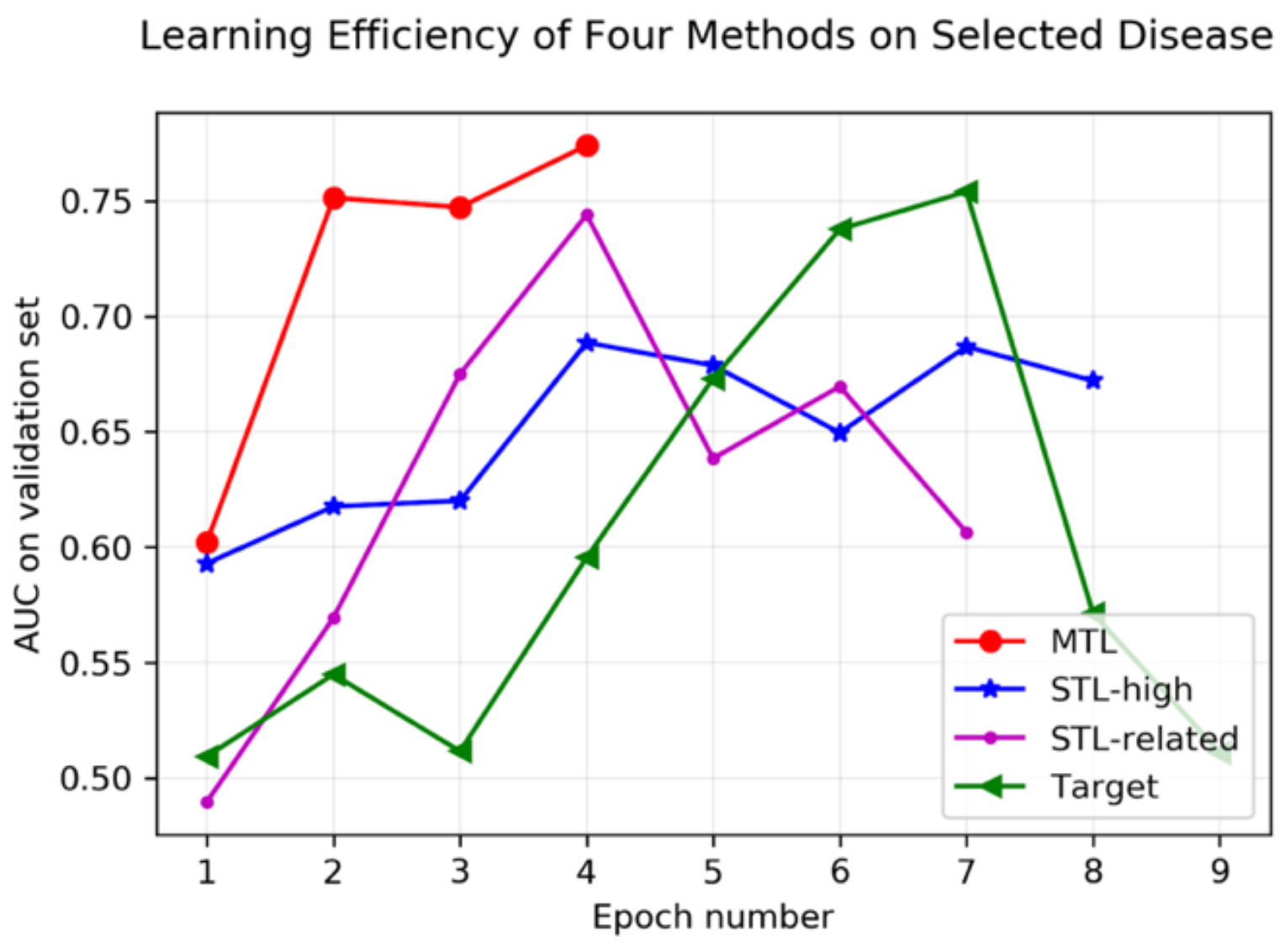

**Figure 4:** Learning efficiency of four methods across training epochs on two selected phenotypes.

[The performances shown here are AUC scores of validation set at each epoch. In both experiments, multi-task pre-training required fewer epochs to achieve the best performance than single-task pre-training or Target without pre-training.]

Because the early stopping mechanism was the same across four methods, our proposed method required less time to converge compared to the other three methods. We hypothesize that this was due to *MTL* having more information from multiple sources compared to the other methods. Furthermore, as the *MTL* pre-trained model was only trained once at the pre-training step, we suggest that the proposed method is more efficient.

## 5. DISCUSSION

In this study, we demonstrated the potential of learning generalized and transferable patient representations from clinical notes with multi-task pre-training and fine-tuning. The pre-trained model used multi-task learning of the five highest-prevalence phenotypes. We further fine-tuned this phenotype-specific pre-trained model to multiple low-prevalence phenotypes that had limited data. The results are promising since the performance on the majority of low-prevalence phenotypes with multi-task pre-trained models was consistently better than the baseline method of only predicting the low-prevalence phenotype using their own data. Further, the multi-task pre-training increases learning efficiency and achieves more robust performance than both baseline methods of single-task pre-training. As we only need to pre-train the multi-task model once, the multi-task pre-trained model is more efficient than the single-task pre-trained model, even when the latter has the most relevant source task associated with the low-prevalence phenotypes (i.e., STL-related). This supports our hypothesis that because the multi-task pre-training phenotypes are related to each other, the multi-task model enables generalizability by sharing a robust representation.

MTL pre-training and fine tuning improved prediction performance for low-prevalence phenotypes. The pre-training uses highest-prevalence phenotypes for each organ system as the source tasks, which allows joint learning organ system-specific models. These organ system-specific pre-trained models are able to learn complex semantics by combining multiple phenotypes that convey different underlying factors and variable context. Therefore, they are capable of handling intricate unseen cases. For simplicity in this work, we only choose five phenotypes for each organ system. In practice, the number of pre-training phenotypes can be much larger than five.

When thinking of how to best adapt the pre-trained model to a given target task, a feature extraction approach would be another option for this architecture. However, unlike the "pre-training and fine-tuning" approach, which adapts all model parameters, the feature extraction approach would obtain a fixed vector from layers of the language model. The representation can be selected from any layer (e.g., last hidden layer, etc.) or any combinations of multiple layers (e.g., weighted sum, concatenation, etc.). Next, the downstream task will be initialized with the extracted vector. There have been multiple studies in open domain fields to compare these two approaches since when the pre-trained language models were introduced [36]. For instance, in Table 7 of Devlin et al. [3], different types of feature-extraction strategies were applied. The performance on the dev sets showed all of them failed to outperform the fine-tuning approach with different sizes of the BERT model. We acknowledge that the feature-based approach certainly has its benefits over the fine-tuning approach, particularly in terms of computational resources. It is much computational faster (in terms of training time) to extract a vector and to use it in many experiments with small models on top of this vector, rather than to restore a deep learning model and to tweak its parameters on the new task.

While a feature extraction approach may be worth evaluating in the future, we assume the fine-tuning is needed to improve the performance for our specific downstream tasks. Because the features extracted with low-dimensional vectors (usually in the hundreds or thousands) are not particularly specialized for the new task, which can be improved by fine-tuning using appropriate hyperparameters specially trained for the new task. When using the feature extraction method, if the model trained on the downstream task was still a deep learning-based model, the experiments might still become computationally expensive. Time-saving comes from the fact that the extracted features only need to be computed once, not because training the target task would save time. Fine-tuning adapts the parameters of the entire model to make them more discriminative for the new task. It also allows us to accommodate a general-purpose representation to various targets. After evaluating trade-offs between time consumption and computational resources with the performance we desire, we still believe fine-tuning is much more appropriate to demonstrate our hypothesis.

When classifying rare diseases, researchers traditionally encounter challenges including low diagnostic rates and limited patient population [37]. Recently, secondary data in the EHR opens perspectives for increasing knowledge of rare diseases. In the meantime, the capability of data-driven approaches to leverage and incorporate multi-source information can be applied to many phenotyping tasks. Therefore, data-driven methods could potentially contribute to overcoming the aforementioned challenges. Some pilot works have tried to harmonize data-driven methods with EHR data to enrich rare disease knowledge [38]. Garcelon et al. [39] applied TF-IDF method to extract concepts from clinical notes in a clinical data warehouse for RETT syndrome. The extracted terms could improve existing phenotypic description. Shen et al. conducted a series of works to enrich rare disease knowledge with data-driven methods, including a graph convolutional network-based recommendation system [40], human phenotype ontology embeddings [41], and association rule mining algorithms [42]. Although, in this work, we attempt to focus on low-prevalence phenotypes (i.e., sample sizes are relatively small) in a specific patient cohort (i.e., MIMIC-III), rather than rare disease in the whole population. We believe these works show the enormous potential to integrate multi-source information and to elucidate insightful understandings of rare diseases. Incorporating these works with our current experimental design and applying them into actual rare disease phenotyping tasks will be one of our future directions.

Despite exploring the benefits of learning generalized and transfer patient representation, we anticipate future work based on the architecture and effectiveness of MTL pre-training. First, this work makes the conventional assumption for predicting phenotypes from clinical notes that the billing codes (ICD-9, ICD-10, etc.) are the actual diagnoses, and can be used for predictive labels of supervised learning. Further work is required to explore additional medical knowledge to augment billing codes. We will incorporate external resources to augment ICD codes to be used as the prediction labels. While developing validated phenotyping algorithms for the 78 low-prevalence phenotypes investigated in this work would be infeasible, we will implement two high-throughput computational phenotyping approaches to enhance the label (i.e., Phecode [43], PheMAP [44]). Further studies are required to prove the validity and reliability of this approach across datasets with precise labels. Additionally, it is necessary to use a hierarchical structure in the neural network to model clinical notes rather than a simple word-level classification with a single, large block of text so as to extract long contiguous and temporal sequences in the note. Hence, another future direction will focus on implementing advanced language models [3,45,46] into hierarchical levels and incorporate them into our learning architecture. In addition, more work on composing different pre-training tasks apart from phenotype predictions is needed, to further exploit the power of pre-trained language representation. Furthermore, the relationship between the number of tasks in MTL pre-training with the target task performance would be instructive to explore. It is assumed that with the number of tasks increasing, the pre-trained model would be enriched with more knowledge about the patient, thus it may be beneficial to the target task prediction. Thus, we will extend our methods to construct supervised learning with different numbers of tasks.

**Limitations**

This study evaluated only a single machine learning model (Si & Roberts [33]) on a single corpus (MIMIC-III [28]). While we would expect the general trend of the results to hold up for different models and different datasets as it is a general-purpose patient phenotyping model, the extent to which these results generalize cannot be empirically determined from this study. A total of 292 model instances were trained in the experiments, so it was deemed infeasible to evaluate several different classes of models or different datasets. However, since the hierarchical RNN model itself is not the core contribution of this work, further studies are necessary to validate this approach is robust across models and datasets. Another limitation is the reliability of ICD codes as labels, since these are known to be poor representations of clinical reality. While developing validated phenotyping algorithms for the 78 low-prevalence phenotypes investigated in this work would be infeasible, future work should investigate experiments with validated phenotyping algorithms for the common phenotypes used in pre-training.

## 6. CONCLUSION

We propose a multi-task pre-training and transfer learning architecture to learn generalized and transferable patient representations from clinical notes. This architecture enables us to obtain a pre-trained model from high-prevalence phenotypes and further fine-tune on low-prevalence phenotypes to achieve good performance even with limited data. The proposed method consistently and oftentimes greatly improves performance and enhances learning efficiency. Ultimately, this pre-training and fine-tuning paradigm may be used to build comprehensive clinical language representations by incorporating the information from heterogeneous clinical sources.

### Author Contributions Statement

YS and KR conceived of the study. EVB aided with the experimental design. YS conducted all experiments and drafted the results. All authors contributed to the final manuscript.

### Conflicts of Interest

None declared

### Acknowledgments

This work was supported by the U.S. National Library of Medicine (NLM) under awards R00LM012104 and R01LM011829; the National Center for Advancing Translational Sciences (NCATS) under awards UL1TR000371 and U01TR002393; the Patient-Centered Outcomes

Research Institute (PCORI) under award ME- 2018C1- 10963; the Cancer Prevention and Research Institute of Texas (CPRIT), under awards RP1500535 and RP170668, and the Reynolds and Reynolds Professorship in Clinical Informatics.

**Appendix Table 1. Low-prevalence Phenotypes with Target as the best performance**

| Disease | ICD-9 | MTL | STL-high | STL-related | Target |
|---|---|---|---|---|---|
| Acute systolic heart failure | 428.21 | 0.9256 | 0.9238 | 0.9124 | 0.9279 |
| Acute diastolic heart failure | 428.31 | 0.8801 | 0.8793 | 0.8793 | 0.9086 |
| Acute on chronic combined systolic and diastolic heart failure | 428.43 | 0.8448 | 0.8173 | 0.833 | 0.8594 |
| Atherosclerosis of aorta | 440.0 | 0.8158 | 0.8345 | 0.844 | 0.8709 |
| Combined systolic and diastolic heart failure, unspecified | 428.40 | 0.8715 | 0.6269 | 0.8559 | 0.8998 |
| Unspecified transient cerebral ischemia | 435.9 | 0.6328 | 0.5479 | 0.5479 | 0.6945 |
| Acute edema of lung, unspecified | 518.4 | 0.8358 | 0.8393 | 0.8332 | 0.8662 |
| Hypertrophy (benign) of prostate | 600.0 | 0.8882 | 0.4491 | 0.9069 | 0.9319 |
| Nephritis and nephropathy, not specified as acute or chronic, with unspecified pathological lesion in kidney | 583.9 | 0.7016 | 0.7093 | 0.7093 | 0.7847 |
| Vascular disorders of kidney | 593.81 | 0.718 | 0.7303 | 0.7963 | 0.9842 |
| Chronic glomerulonephritis in diseases classified elsewhere | 582.81 | 0.2513 | 0.0676 | 0.1894 | 0.5202 |

**Appendix table 2. Descriptive statistics of low-prevalence phenotypes in three organ systems with the cases for the test sets.**

1. Circulatory

| Disease Name | ICD-9 | # cases | # cases in test set |
|---|---|---|---|
| Acute systolic heart failure | 428.21 | 492 | 37 |
| Coronary atherosclerosis of autologous vein bypass graft | 414.02 | 474 | 33 |
| Other late effects of cerebrovascular disease | 438.89 | 465 | 30 |
| Benign essential hypertension | 401.1 | 454 | 30 |
| Late effects of cerebrovascular disease, hemiplegia affecting unspecified side | 438.20 | 437 | 27 |
| Acute diastolic heart failure | 428.31 | 432 | 26 |
| Systolic heart failure, unspecified | 428.20 | 416 | 31 |
| Subdural hemorrhage | 432.1 | 392 | 24 |
| Sinoatrial node dysfunction | 427.81 | 389 | 25 |
| Acute myocardial infarction of unspecified site, initial episode of care | 410.91 | 354 | 28 |
| Atherosclerosis of native arteries of the extremities with gangrene | 440.24 | 327 | 24 |
| Acute on chronic combined systolic and diastolic heart failure | 428.43 | 327 | 21 |
| Chronic total occlusion of coronary artery | 414.2 | 292 | 27 |
| Atherosclerosis of aorta | 440.0 | 283 | 20 |
| Sub-endocardial infarction, subsequent episode of care | 410.72 | 279 | 19 |
| Atherosclerosis of native arteries of the extremities with ulceration | 440.23 | 264 | 17 |
| Atherosclerosis of native arteries of the extremities with intermittent claudication | 440.21 | 257 | 19 |
| Late effects of cerebrovascular disease, aphasia | 438.11 | 240 | 22 |
| Atherosclerosis of renal artery | 440.1 | 233 | 17 |
| Iatrogenic hypotension | 458.2 | 233 | 15 |
| Cerebral atherosclerosis | 437.0 | 196 | 12 |
| Hypertrophic cardiomyopathy | 425.1 | 183 | 14 |
| Chronic combined systolic and diastolic heart failure | 428.42 | 179 | 10 |
| Malignant essential hypertension | 401.0 | 172 | 10 |
| Paroxysmal supraventricular tachycardia | 427.0 | 161 | 11 |
| Acute myocardial infarction of anterolateral wall, initial episode of care | 410.01 | 142 | 9 |
| Hypertensive chronic kidney disease, benign, with chronic kidney disease stage I through stage IV, or unspecified | 403.10 | 122 | 8 |
| Combined systolic and diastolic heart failure, unspecified | 428.40 | 110 | 5 |
| Unspecified transient cerebral ischemia | 435.9 | 96 | 6 |
| Atherosclerosis of native arteries of the extremities with rest pain | 440.22 | 88 | 5 |
| Abdominal aneurysm, ruptured | 441.3 | 76 | 5 |
| Acute combined systolic and diastolic heart failure | 428.41 | 73 | 5 |
| Unspecified late effects of cerebrovascular disease | 438.9 | 69 | 7 |
| Unspecified cerebrovascular disease | 437.9 | 64 | 6 |
| Atherosclerosis of other specified arteries | 440.8 | 63 | 4 |
| Cerebral thrombosis with cerebral infarction | 434.01 | 60 | 7 |
| Secondary cardiomyopathy, unspecified | 425.9 | 53 | 3 |
| Acute myocardial infarction of unspecified site, subsequent episode of care | 410.92 | 53 | 3 |

2. Respiratory

| Disease Name | ICD-9 | # cases | # cases in test set |
|---|---|---|---|
| Post-inflammatory pulmonary fibrosis | 515 | 544 | 35 |
| Pneumonia due to Pseudomonas | 482.1 | 430 | 31 |
| Chronic respiratory failure | 518.83 | 331 | 29 |
| Acute edema of lung, unspecified | 518.4 | 305 | 20 |
| Chronic obstructive asthma with (acute) exacerbation | 493.22 | 299 | 25 |
| Pneumonia due to other gram-negative bacteria | 482.83 | 264 | 18 |
| Bacterial pneumonia, unspecified | 482.9 | 227 | 19 |
| Pneumonia due to Klebsiella pneumoniae | 482.0 | 226 | 15 |
| Pneumococcal pneumonia [Streptococcus pneumoniae pneumonia] | 481 | 194 | 17 |
| Bronchiectasis without acute exacerbation | 494.0 | 191 | 13 |
| Empyema without mention of fistula | 510.9 | 190 | 15 |
| Methicillin resistant pneumonia due to Staphylococcus aureus | 482.42 | 162 | 10 |
| Pulmonary congestion and hypostasis | 514 | 155 | 9 |
| Unspecified sinusitis (chronic) | 473.9 | 149 | 7 |
| Edema of larynx | 478.6 | 145 | 10 |
| Malignant pleural effusion | 511.81 | 132 | 8 |
| Acute bronchitis | 466.0 | 126 | 8 |
| Asbestosis | 501 | 116 | 9 |
| Acute upper respiratory infections of unspecified site | 465.9 | 96 | 6 |
| Abscess of lung | 513.0 | 86 | 9 |
| Unilateral paralysis of vocal cords or larynx, partial | 478.31 | 74 | 4 |
| Empyema with fistula | 510.0 | 72 | 3 |
| Stenosis of larynx | 478.74 | 61 | 4 |

3. Genitourinary

| Disease Name | ICD-9 | # cases | # cases in test set |
|---|---|---|---|
| Hematuria | 599.7 | 509 | 34 |
| Hydronephrosis | 591 | 413 | 29 |
| Chronic kidney disease, Stage IV (severe) | 585.4 | 334 | 25 |
| Hypertrophy (benign) of prostate with urinary obstruction and other lower urinary tract symptoms (LUTS) | 600.01 | 314 | 27 |
| Neurogenic bladder NOS | 596.54 | 225 | 18 |
| Hematuria, unspecified | 599.70 | 216 | 16 |
| Calculus of kidney | 592.0 | 206 | 12 |
| Gross hematuria | 599.71 | 181 | 12 |
| Secondary hyperparathyroidism (of renal origin) | 588.81 | 169 | 8 |
| Acute pyelonephritis without lesion of renal medullary necrosis | 590.10 | 132 | 13 |
| Calculus of ureter | 592.1 | 111 | 7 |
| Cyst of kidney, acquired | 593.2 | 107 | 8 |
| Hypertrophy (benign) of prostate | 600.0 | 92 | 6 |
| Pyelonephritis, unspecified | 590.80 | 84 | 8 |
| Nephritis and nephropathy, not specified as acute or chronic, with unspecified pathological lesion in kidney | 583.9 | 84 | 6 |
| Vascular disorders of kidney | 593.81 | 83 | 5 |
| Chronic glomerulonephritis in diseases classified elsewhere | 582.81 | 67 | 3 |